\documentclass[twoside]{article}
\usepackage[accepted]{aistats2015}
\usepackage[round]{natbib}
\usepackage{hyperref}
\usepackage{url}
\usepackage{amsmath}
\usepackage{amsthm}
\usepackage{amssymb}
\usepackage{algorithm}
\usepackage{algpseudocode}
\usepackage{graphicx}
\graphicspath{{images/}}
\usepackage{subfigure}
\newtheorem{theorem}{Theorem}

\newtheorem{prop}{Proposition}
\newtheorem{defn}{Definition}

\usepackage{flushend}

\begin{document}

%

%

\twocolumn[

\aistatstitle{Learning Planar Ising Models}
\aistatsauthor{Jason K. Johnson \And Diane Oyen \And Michael Chertkov \And Praneeth Netrapalli}
\aistatsaddress{Numerica \\ Ft. Collins, CO, USA \And Los Alamos National Lab \\ Los Alamos, NM, USA 
	\And Los Alamos National Lab \\ Los Alamos, NM, USA \And Microsoft Research \\ Cambridge, MA, USA}]

\begin{abstract} 
Inference and learning of graphical models are both well-studied problems in
statistics and machine learning that have found many applications in science and
engineering. However, exact inference is intractable in general graphical
models, which suggests the problem of seeking the best approximation to a
collection of random variables within some tractable family of graphical models.
 In this paper, we focus on the class of planar Ising models, for
which exact inference is tractable using techniques of statistical physics.
Based on these techniques and recent methods for planarity
testing and planar embedding,
we propose a simple greedy
algorithm for learning the best planar Ising model to approximate an arbitrary
collection of binary random variables (possibly from sample data). Given
the set of all pairwise correlations among variables, we select a planar graph and
optimal planar Ising model defined on this graph to best approximate that
set of correlations.  We demonstrate our method in simulations and for
the application of modeling senate voting records.
\end{abstract}

\section{Introduction}\label{sec:intro}

Graphical models are widely used to represent the statistical relations among a
set of random variables \citep{Lauritzen,MacKay}. Nodes of the graph correspond to random variables and
edges of the graph represent statistical interactions among the variables. The
problems of inference and learning on graphical models arise in many
practical applications. The problem of inference is to deduce certain
statistical properties (such as marginal probabilities, modes etc.) of a given
set of random variables whose graphical model is known. It has wide applications
in areas such as error correcting codes, statistical physics and so on.
The problem of learning on the other hand is to deduce the graphical
model of a set of random variables given statistics (possibly from samples) of
the random variables. Learning is also a widely encountered problem in areas
such as biology, anthropology and so on. 

The \emph{Ising model}, a class of binary-variable graphical models with pairwise interactions,
has been studied by physicists as a simple model
of order-disorder transitions in magnetic materials \citep{onsager}. 
Remarkably, it was found that in the special case of an Ising model with
zero-mean $\{-1,+1\}$ binary random variables and pairwise interactions defined
on a planar graph, calculation of the partition function (which is closely tied
to inference) is tractable, essentially reducing to calculation of a matrix
determinant \citep{kacward,sherman,kasteleyn,fisher}.
These methods have been used in machine learning \citep{schraudolph,globerson}.

We address the problem of approximating a collection of binary
random variables (given their pairwise marginal distributions) by a zero-mean
planar Ising model.  We also consider the related problem of selecting a
non-zero mean Ising model defined on an outer-planar graph (these models are
also tractable, being essentially equivalent to a zero-field model on a related
planar graph).

There has been a great deal of work on learning graphical models. Much of these
have focused on learning over the class of thin graphical models
\citep{deshpande,bach,karger,shahaf} for which inference is tractable by converting the
model to a junction tree.  The simplest case of this is learning tree
models (treewidth one graphs) for which it is tractable to find the best tree
model by reduction to a max-weight spanning tree problem \citep{chowliu}. 
However, the problem of finding the best bounded-treewidth model is NP-hard for
treewidths greater than two \citep{karger}, and so heuristic methods are used to
select the graph structure \citep{deshpande,karger}.  Another popular method is to use convex optimization of
the log-likelihood penalized by $\ell_1$ norm of parameters of the graphical
model so as to promote sparsity \citep{banerjee,lee}. To go beyond low-treewidth
graphs, such methods either focus on Gaussian graphical models or adopt a
tractable approximation of the likelihood.  Other methods learn only the
graph structure itself \citep{ravikumar,abbeel} and are often able to demonstrate
asymptotic correctness of this estimate under appropriate conditions.  

In contrast to existing approaches, this paper explores planarity as an alternative restriction on the model class to both make learning tractable and to offer a qualitatively different graph topology in which the number of edges learned is linear in the number of variables.



\section{Preliminaries}\label{sec:prelim}

In this section, we develop our notation and briefly review the necessary
background theory. 

\subsection{Divergence and Likelihood}\label{subsec:entropy}

Suppose we want to calculate how well a probability distribution $Q$
approximates another probability distribution $P$ (on the same sample space
$\chi$). For any two probability distributions $P$ and $Q$ on some sample space
$\chi$, we denote by $D(P,Q)$ the \emph{Kullback-Leibler divergence} (or
\emph{relative entropy}) between $P$ and $Q$ as 
$D(P,Q) =  \sum_{x \in \chi} P(x) \log \frac{P(x)}{Q(x)}$.
The \emph{log-likelihood function} is defined as 
$LL(P,Q) =  \sum_{x \in \chi} P(x) \log Q(x)$.
The probability distribution in a family $\mathcal{F}$ that maximizes the
log-likelihood of a probability distribution $P$ is called the
\emph{maximum-likelihood estimate} of $P$ in $\mathcal{F}$, and this is
equivalent to the \emph{minimum-divergence projection} of $P$ to $\mathcal{F}$, so that
$P_\mathcal{F} 
=  \operatorname*{arg\,max}_{Q \in \mathcal{F}} LL(P,Q) 
=  \operatorname*{arg\,min}_{Q \in \mathcal{F}} D(P,Q)$.

\subsection{Graphical Models and The Ising Model}
\label{subsec:brush}

We will be dealing with binary random variables throughout
the paper. We write $P(x)$ to denote the probability distribution of a
collection of random variables $x=(x_1,\dots,x_n)$. Unless otherwise stated, we
work with undirected graphs $G=(V,E)$ with vertex (or node) set $V$ and edges
$\{i,j\} \in E \subset {V \choose 2}$. For vertices $i,j \in V$ we write $G+ij$
to denote the graph $(V,E \cup \{i,j\})$.
A \emph{pairwise graphical model} is a probability distribution $P(x) =
P(x_1,\dots,x_n)$ that is defined on a graph $G = (V,E)$ with vertices $V =
\{1,..,n\}$ as
\begin{equation} \label{eq:graphical_model}
\begin{split}
	P(x) & \propto \quad \prod_{i \in V} \psi_i(x_i) \prod_{\{i,j\} \in E} \psi_{ij}(x_i,x_j) \\
		& \propto \quad \exp \biggr\{ \sum_{i \in V} f_i(x_i) + \sum_{\{i,j\} \in E} f_{ij}(x_i,x_j) \biggr\}
\end{split}
\end{equation}
where $\psi_i,\psi_{ij} \geq 0$ are non-negative node and edge compatibility
functions. For positive $\psi$'s, we may also represent $P(x)$ as a Gibbs
distribution with potentials $f_i = \log \psi_i$ and $f_{ij} = \log \psi_{ij}$. 

\begin{defn}
An \emph{Ising model} on binary random variables $x = (x_1,\dots,x_n)$ and graph
$G=(V,E)$ is the probability distribution defined by 
\begin{align*}
	P(x) &= \frac{1}{Z(\theta)} \exp\biggr\{\sum_{i \in V} \theta_i x_i + \!\!\!\sum_{\{i,j\} \in E}\!\!\theta_{ij}x_ix_j\biggr\}, \\
	Z(\theta) &= \sum_x \exp\biggr\{\sum_{i \in V} \theta_i x_i + \!\!\!\sum_{\{i,j\} \in E}\!\!\theta_{ij}x_ix_j\biggr\},
\end{align*}
where $x_i\in \{-1,1\}$. The \emph{partition function} $Z(\theta)$ serves to normalize the
probability distribution.
\end{defn}

Formally, this defines an \emph{exponential family} $P_\theta(x) = \exp\{ \theta^T \phi(x) - \Phi(\theta)\}$
\citep{barndorff,wainwright} based on sufficient statistics $(\phi_i(x)=x_i,
i \in V)$ and $(\phi_{ij}(x) = x_i x_j, \{i,j\} \in E)$, parameters $(\theta_i, i \in V)$ 
and $(\theta_{ij}, \{i,j\} \in E)$ and moment parameters 
$(\mu_i = \mathbb{E}[x_i], i \in V)$ and $(\mu_{ij} = \mathbb{E}[x_i x_j], \{i,j\} \in E)$.
The function $\Phi(\theta) = \log Z(\theta)$ is a convex function
of $\theta$ and has the moment generating properties: $\nabla \Phi(\theta) = \mathbb{E}_\theta[\phi(x)] = \mu$
and $\nabla^2 \Phi(\theta) = \mathbb{E}_\theta[(\phi(x) - \mu)(\phi(x) - \mu)^T]$.

In fact, any pairwise graphical model among binary variables can be represented as an Ising 
model:
\begin{align*}
	\theta_i &= \tfrac{1}{2} \sum_{x_i} x_i f_i(x_i) 
		+ \tfrac{1}{4} \sum_{\{i,j\} \in E} \sum_{x_i,x_j} x_i f_{ij}(x_i,x_j),\\
	\theta_{ij} &= \tfrac{1}{4} \sum_{x_i,x_j} x_i x_j f_{ij}(x_i,x_j) . 
\end{align*}
The moments can be computed as: $\mu_i = \sum_{x_i} x_i
P(x_i)$ and $\mu_{ij} = \sum_{x_i,x_j} x_i x_j P(x_i,x_j)$. Inversely, the marginals are
computed by:
\begin{align*}
P(x_i) &= \tfrac{1}{2} (1 + \mu_i x_i), \\
P(x_i,x_j) &= \tfrac{1}{4} (1 + \mu_i x_i + \mu_j x_j + \mu_{ij} x_i x_j) .
\end{align*}

An Ising model is said to be \emph{zero-field} if $\theta_i = 0$ for all $i \in
V$. It is \emph{zero-mean} if $\mu_i = 0$ ($P(x_i = \pm 1) = \tfrac{1}{2}$) for
all $i \in V$.
The Ising model is zero-field if and only if it is
zero-mean. Although the zero-field assumption appears very restrictive, a
general Ising model can be represented as a zero-field model by adding one
auxiliary variable node connected to every other node of the graph \citep{globerson}. The
parameters and moments of the two models are then related as follows:
\begin{prop}\label{prop:zeromeantononzero}
Consider the Ising model on $G=(V,E)$ with $V = \{1,\dots,n\}$, parameters
$\{\theta_i\}$ and $\{\theta_{ij}\}$, moments $\{\mu_i\}$ and $\{\mu_{ij}\}$ and
partition function $Z$. Let $\widehat{G} = (\widehat{V},\widehat{E})$ denote the extended
graph based on nodes $\widehat{V} = V \cup \{n+1\}$ with edges $\widehat{E} = E \cup
\{\{i,n+1\}, i \in V \})$. We define a zero-field Ising model on $\widehat{G}$ with
parameters $\{\widehat\theta_{ij}\}$, moments $\{\widehat\mu_{ij}\}$ and partition
function $\widehat{Z}$. If we set the parameters according to
\begin{displaymath}
\widehat\theta_{ij} =
\left\{
\begin{array}{ll}
 \theta_i ~\mbox{if}~ j=n+1\\
 \theta_{ij} ~\mbox{otherwise}
\end{array}
\right.
\end{displaymath}
then $\widehat{Z} = 2 Z$ and 
\begin{displaymath}
\widehat\mu_{ij} =
\left\{
\begin{array}{ll}
 \mu_i ~\mbox{if}~ j=n+1\\
 \mu_{ij} ~\mbox{otherwise}
\end{array}
\right.
\end{displaymath}
\end{prop}
Thus, inference on the corresponding zero-field Ising model on the extended
graph $\widehat{G}$ is equivalent to inference on the (non-zero-field)
Ising model defined on $G$. Proof given in the Supplement.

\subsection{Inference for Planar Ising Models}\label{subsec:partfunc_planarising}

A graph is \emph{planar} if it may be embedded in the plane without any edge crossings. 
It is known that any planar graph can be embedded such that all edges 
are drawn as straight lines. The motivation for our paper is the following 
result on tractability of inference for the \emph{planar zero-field Ising model}.

\begin{theorem}\label{thm:partfunc_planarising}
\citep{kacward,sherman,loebl}
Let $G$ be a planar graph with specified straight-line embedding in the plane
and let $\phi_{ijk} \in [-\pi,+\pi]$ denote the clockwise rotation between 
the directed edges $(i,j)$ and $(j,k)$. We define
the matrix $W \in \mathbb{C}^{2 |E| \times 2 |E|}$ indexed by directed edges of
the graph as follows: $W = A D$ where $D$ is the diagonal matrix with $D_{ij,ij}
= \tanh \theta_{ij} \triangleq w_{ij}$ and
\begin{displaymath}
A_{ij,kl} = 
\left\{ 
\begin{array}{ll}
\exp(\tfrac{1}{2} \sqrt{-1} \phi_{ijl}), &  j=k ~\mbox{and}~ i \neq l \\
0, & \mbox{otherwise}
\end{array}
\right.
\end{displaymath}
Then, the partition function of the zero-field planar Ising model is given by the Kac-Ward determinant formula:
\begin{displaymath}
 Z = 2^n \biggr(\prod_{\{i,j\}\in E} \cosh \theta_{ij}\biggr) \det(I-W)^{\frac{1}{2}}
\end{displaymath}
\end{theorem}
Another related method for computing the Ising model partition function is 
based on counting perfect matchings of planar graphs \citep{kasteleyn,fisher}.
Thus, calculating the partition function reduces to calculating the
determinant of a matrix; therefore, using the
generalized nested dissection algorithm to exploit sparsity of the matrix, the
complexity of these calculations is $O(n^{3/2})$ \citep{liptonrose,liptontarjan,galluccio}.
Thus, inference of the zero-field planar Ising model is tractable and scales well with
problem size.

The gradient and Hessian of the log-partition function
$\Phi(\theta) = \log Z(\theta)$ can also be calculated efficiently from the Kac-Ward
determinant formula. Derivatives of $\Phi(\theta)$ recover the
moment parameters of the exponential family model as 
$\nabla \Phi(\theta) = \mathbb{E}_\theta[\phi] = \mu$  \citep{barndorff,wainwright}.
Thus, inference of moments (and node and edge marginals) are tractable
for the zero-field planar Ising model.

\begin{prop}\label{prop:gradienthessiancalc} 
Let $\mu = \nabla \Phi(\theta)$, $H = \nabla^2 \Phi(\theta)$. Let $S =
(I-W)^{-1} A$ and $T = (I+P) (S \circ S^T) (I+P^T)$ where $A$ and $W$ are defined as in
Theorem 1, $\circ$ denotes the element-wise product and $P$ is the permutation matrix
swapping indices of directed edges $(i,j)$ and $(j,i)$. Then,
\begin{displaymath}
\begin{split}
	\mu_{ij} &=  w_{ij} - \tfrac{1}{2} (1-w^2_{ij}) (S_{ij,ij} + S_{ji,ji}) \\
	H_{ij,kl} &= 
		\left\{
		\begin{array}{ll}
			1-\mu_{ij}^2, & ij = kl  \\
			- \tfrac{1}{2} (1-w^2_{ij}) T_{ij,kl} (1-w^2_{kl}), & ~\mbox{otherwise}
		\end{array}
		\right.
\end{split}
\end{displaymath}
\end{prop}
 
Calculating the full matrix $S$ requires $O(n^3)$ calculations. However,
to compute just the moments $\mu$ only the diagonal elements of $S$ are needed. 
Then, using the generalized nested dissection method, inference of moments
(edge-wise marginals) of the zero-field Ising model can be achieved with
complexity $O(n^{3/2})$.
Computing the full Hessian is more expensive, requiring $O(n^3)$ calculations.

\paragraph{Inference for Outer-Planar Graphical Models}

We emphasize that the above calculations require both a planar graph $G$ and a
zero-field Ising model. Using the graphical transformation of Proposition 1,
the latter zero-field condition may be relaxed but at the expense of adding an
auxiliary node connected to all the other nodes. In general planar graphs $G$,
the new graph $\widehat{G}$ may not be planar and hence may not admit tractable
inference calculations. However, for the subset of planar graphs where this
transformation does preserve planarity inference is still tractable.

\begin{defn} A graph $G$ is said to be \emph{outer-planar} if there exists an
embedding of $G$ in the plane where all the nodes are on the outer face.
\end{defn}

In other words, the graph $G$ is outer-planar if the extended graph $\widehat{G}$
(defined by Proposition 1) is planar. Then, from Proposition 1 and Theorem 1 it
follows that:

\begin{prop} \citep{globerson} The partition function and moments of any outer-planar Ising
graphical model (not necessarily zero-field) can be calculated efficiently. 
Hence, inference is tractable for any binary-variable graphical model with
pairwise interactions defined on an outer-planar graph. \end{prop}

This motivates the problem of learning outer-planar graphical models
for a collection of (possibly non-zero mean) binary random variables.

\section{Learning Planar Ising Models}\label{sec:main}

This section addresses the main goals of the paper, which are two-fold:
\vspace{-6pt}
\begin{enumerate}
\item Solving for the maximum-likelihood Ising model on a given planar graph
to best approximate a collection of zero-mean random variables.
\item How to select (heuristically) the planar graph to obtain the best
approximation.
\end{enumerate}
\vspace{-6pt}
We address these problems in the following 
two subsections. The solution of the first problem is an integral
part of our approach to the second. Both solutions are easily
adapted to the context of learning outer-planar graphical models of
(possibly non-zero mean) binary random variables.

\subsection{Maximum-Likelihood Parameter Estimation}\label{subsec:convexopt}

Maximum-likelihood estimation over an exponential
family is a convex optimization problem based on the
log-partition function $\Phi(\theta)$. In the case of the zero-field Ising model
defined on a given planar graph it is tractable to compute $\Phi(\theta)$ via a
matrix determinant described in Theorem 1. Thus, we obtain an unconstrained,
tractable, convex optimization problem for the maximum-likelihood zero-field
Ising model on the planar graph $G$ to best approximate a probability
distribution $P(x)$:
\begin{align*}
&\max_{\theta} \{ \mu^T \theta - \Phi(\theta) \} = \\
&\max_{\theta \in \mathbb{R}^{|E|}}
	\!\biggl\{ \!\sum_{ij} (\mu_{ij} \theta_{ij} - \log\cosh \theta_{ij}) - \tfrac{1}{2} \log\det (I-W(\theta)) \!\biggr\}
\end{align*}
Here, $\mu_{ij} = \mathbb{E}_P[x_i x_j]$ for all edges $\{i,j\} \in G$ and the
matrix $W(\theta)$ is as defined in Theorem 1. If $P$ represents the empirical
distribution of a set of independent identically-distributed (iid) samples
$\{x^{(s)}, s=1,\dots,S\}$ then $\{\mu_{ij}\}$ are the corresponding empirical
moments $\mu_{ij} = \frac{1}{S} \sum_s x^{(s)}_i x^{(s)}_j$.

\paragraph{Newton's Method} 

We solve this unconstrained convex optimization problem using Newton's method
with step-size chosen by back-tracking line search \citep{boyd}. This produces a
sequence of estimates $\theta^{(t)}$ calculated as follows:
\begin{displaymath}
\theta^{(t+1)} = \theta^{(t)} + \lambda_t H(\theta^{(t)})^{-1} ( \mu(\theta^{(t)}) - \mu)
\end{displaymath}
where $\mu(\theta^{(t)})$ and $H(\theta^{(t)})$ are calculated using Proposition 2 and
$\lambda_t \in (0,1]$ is a step-size parameter chosen by backtracking line
search (see \cite{boyd}: Chapter 9, Section 2 for details). The per iteration complexity of this
optimization is $O(n^3)$ using explicit computation of the Hessian at each
iteration. This complexity can be offset somewhat by only re-computing the
Hessian a few times (reusing the same Hessian for a number of iterations), to
take advantage of the fact that the gradient computation only requires
$O(n^\frac{3}{2})$ calculations. As Newton's method has quadratic
convergence, the number of iterations required to achieve a high-accuracy
solution is typically 8-16 iterations (essentially independent of problem size).
 We estimate the computational complexity of solving this convex optimization
problem as roughly $O(n^3)$.

\subsection{Greedy Planar Graph Selection}\label{subsec:problem}

We now consider the problem of selection of the planar graph $G$ to best
approximate a probability distribution $P(x)$ with pairwise moments $\mu_{ij} =
\mathbb{E}_P[x_i x_j]$ given for all $i,j \in V$. Formally, we seek the planar
graph that maximizes the log-likelihood (minimizes the divergence) relative to
$P$:
\begin{displaymath}
\widehat{G} 
= \operatorname*{arg\,max}_{G \in \mathcal{P}_V} LL(P,P_G) 
= \operatorname*{arg\,max}_{G \in \mathcal{P}_V} \max_{Q \in \mathcal{F}_G} LL(P,Q)
\end{displaymath}
where $\mathcal{P}_V$ is the set of planar graphs on the vertex set $V$,
$\mathcal{F}_G$ denotes the family of zero-field Ising models defined on graph
$G$ and $P_G = \operatorname*{arg\,max}_{Q \in \mathcal{F}_G} LL(P,Q)$ is the maximum-likelihood
(minimum-divergence) approximation to $P$ over this family.

We obtain a heuristic solution to this graph selection problem using the
following greedy edge-selection procedure. The input to the algorithm is a
probability distribution $P$ (which could be empirical) on $n$ binary $\{-1,1\}$
random variables. In fact, it is sufficient to summarize $P$ by its pairwise
correlations $\mu_{ij} = \mathbb{E}_P[x_i x_j]$ on all pairs $i,j \in V$. The
output is a maximal planar graph $G$ and the maximum-likelihood approximation
$\theta_G$ to $P$ in the family of zero-field Ising models defined on this
graph.
A maximal planar graph is a planar graph for which no new edge can be 
added that would maintain planarity.


\begin{algorithm*}
\caption{GreedyPlanarGraphSelect($P$)}
\label{GLPI}
\begin{algorithmic}[1]
 \State $G = \emptyset, \theta_G = 0$
 \For {$k = 1:3n-6$} 
 	\Comment {Add edges until maximal planar graph reached}
   \State $\Delta = \left\{ \{i,j\} \subset V |  \{i,j\} \notin G, G+ij \in \mathcal{P}_V \right\}$    
       \Comment {Set of candidate edges that preserve planarity}
   \State $\tilde\mu_\Delta = \{ \tilde\mu_{ij} = \mathbb{E}_{\theta_G}[x_i x_j], \{i,j\} \in \Delta \}$ 
   	\Comment {Compute pairwise correlations}
   \State $G \leftarrow G \cup \displaystyle \operatorname*{arg\,max}_{e \in \Delta} D(P_e, \tilde{P}_e)$
   	\Comment {Select edge that maximizes improvement in log-likelihood}
   \State $\theta_G = \mbox{PlanarIsing}(G,P$)
   	\Comment {Compute maximum-likelihood parameters for $G$}
\EndFor
\end{algorithmic}
\end{algorithm*}

The algorithm starts with an empty graph and then sequentially adds edges to the
graph one at a time so as to (heuristically) increase the log-likelihood
(decrease the divergence) relative to $P$ as much as possible at each step. 
Here is a more detailed description of the algorithm along with estimates of the
computational complexity of each step: 
\begin{itemize}
\vspace{-6pt}

\item \emph{Line 3.} First, we enumerate the set $\Delta$ of all edges one
might add (individually) to the graph while preserving planarity. This is
accomplished by an $O(n^3)$ algorithm in which we iterate over all pairs
$\{i,j\} \not\in G$ and for each such pair we form the graph $G + ij$ and
test planarity of this graph using known $O(n)$ algorithms \citep{chrobak}.

\item \emph{Line 4.} Next, we perform tractable inference calculations with
respect to the Ising model on $G$ to calculate the pairwise correlations
$\tilde{\mu}_{ij}$ for all pairs $\{i,j\} \in \Delta$. This is accomplished
using $O(n^{3/2})$ inference calculations on augmented versions of the graph
$G$. For each inference calculation we add as many edges to $G$ from $\Delta$
as possible (setting $\theta = 0$ on these edges) while preserving planarity and
then calculate all the edge-wise moments of this graph using Proposition 2
(including the zero-edges). This requires at most $O(n)$ iterations to cover
all pairs of $\Delta$, so the worst-case complexity to compute all required
pairwise moments is $O(n^{5/2})$.

\item \emph{Line 5.} Once we have these moments, which specify the corresponding
pairwise marginals of the current Ising model, we compare these moments
(pairwise marginals) to those of the input distribution $P$ by evaluating the
pairwise KL-divergence between the Ising model and $P$. As seen by the
following proposition, this gives us a lower-bound on the improvement obtained
by adding edge $\{i,j\}$ (see Supplement for proof):
\begin{prop}\label{prop:lboundKLD}
Let $P_G$ and $P_{G+ij}$ be projections of $P$ on $G$ and $G+ij$ respectively. 
Then,
\begin{displaymath}
D(P, P_{G}) - D(P, P_{G+ij}) \geq D\left( P(x_i,x_j) , P_G(x_i,x_j) \right) 
\end{displaymath}
where $P(x_i,x_j)$ and $P_G(x_i,x_j)$ represent the marginal distributions on $x_i,x_j$ of 
probabilities $P$ and $P_G$ respectively.
\end{prop}
Thus, we greedily select the next edge $\{i,j\}$ to add so as to maximize this
lower-bound on the improvement measured by the increase on log-likelihood (this
being equal to the decrease in KL-divergence).

\item \emph{Line 6.} Finally, we calculate the new maximum-likelihood parameters
$\theta_G$ on the new graph $G \leftarrow G +ij$. This involves solving
the convex optimization problem discussed in the preceding subsection, which
requires $O(n^3)$ complexity. This step is necessary in order to subsequently
calculate the pairwise moments $\tilde\mu$ which guide further edge-selection
steps, and also to provide the final estimate.

\end{itemize}
\vspace{-6pt}

We continue adding one edge at a time until a maximal planar graph (with $3n-6$
edges) is obtained. Thus, the total complexity of our greedy algorithm for
planar graph selection is $O(n^4)$.

\paragraph{Non-Maximal Planar Graphs}

Since adding an edge always improves the log-likelihood, the
greedy algorithm always outputs a maximal planar graph. However, this might
lead to over-fitting of the data
especially when the input probability distribution is
an empirical distribution. Note that at $3n-6$ edges,
the maximal planar graph is sparse and our empirical work indicates that over-fitting is often not an issue. 
In the case that
over-fitting is a concern, we could terminate the algorithm when
adding an edge to the graph would only improve the log-likelihood by less
than some threshold $\gamma$.
An experimental search can be performed for a
suitable value of this threshold (e.g. so as to minimize some estimate of the
generalization, such as in cross validation methods \citep{zhang}). Or, one
could use some heuristic value for $\gamma$ based on the number of samples such
as Akaike's information criterion (AIC) or Shwarz's Bayesian information
criterion (BIC) \citep{akaike,schwarz}.

\paragraph{Outer-Planar Graphs and Non-Zero Means}

The greedy algorithm returns a zero-field Ising model (which has zero mean for
all the random variables) defined on a planar graph. If the actual random
variables are non-zero mean, this may not be desirable. For this case we may
prefer to exactly model the means of each random variable but still retain
tractability by restricting the greedy learning algorithm to select outer-planar
graphs. This model faithfully represents the marginals of each random variable
but at the cost of modeling fewer pairwise interactions among the variables.  

This is equivalent to the following procedure. First, given the sample moments
$\{\mu_i\}$ and $\{\mu_{ij}\}$ we convert these to an equivalent set of
zero-mean moments $\widehat\mu$ on the extended vertex set $\widehat{V} = V \cup
\{n+1\}$ according to Proposition 1. Then, we select a zero-mean planar Ising
model for these moments using our greedy algorithm. However, to fit the means
of each of the original $n$ variables, we initialize this graph to include all
the edges $\{i,n+1\}$ for all $i \in V$ 
(requiring that these are present in our final estimate of the graph $\widehat{G}$).
After this initialization step, we use
the same greedy edge-selection procedure as before. This yields the graph
$\widehat{G}$ and parameters $\theta_{\widehat{G}}$. Lastly, we convert back to a
(non-zero field) Ising model on the subgraph of $\widehat{G}$ defined on nodes $V$,
as prescribed by Proposition 1. The resulting graph $G$ and parameters
$\theta_G$ is our heuristic solution for the maximum-likelihood outer-planar 
Ising model.  

We remark that it is not essential to choose between the
zero-field planar Ising model and the outer-planar Ising model.  
The greedy algorithm may instead select something in between---a partial outer-planar
Ising model where only nodes of the outer-face are allowed to have non-zero
means. This is accomplished simply by omitting the initialization step of
adding edges $\{i,n+1\}$ for all $i \in V$.

\section{Experiments}

\begin{figure*}[tb]
\centering
\vspace{-15pt}
\subfigure[$7 \times 7$ grid]{
	\includegraphics[width=2.1cm]{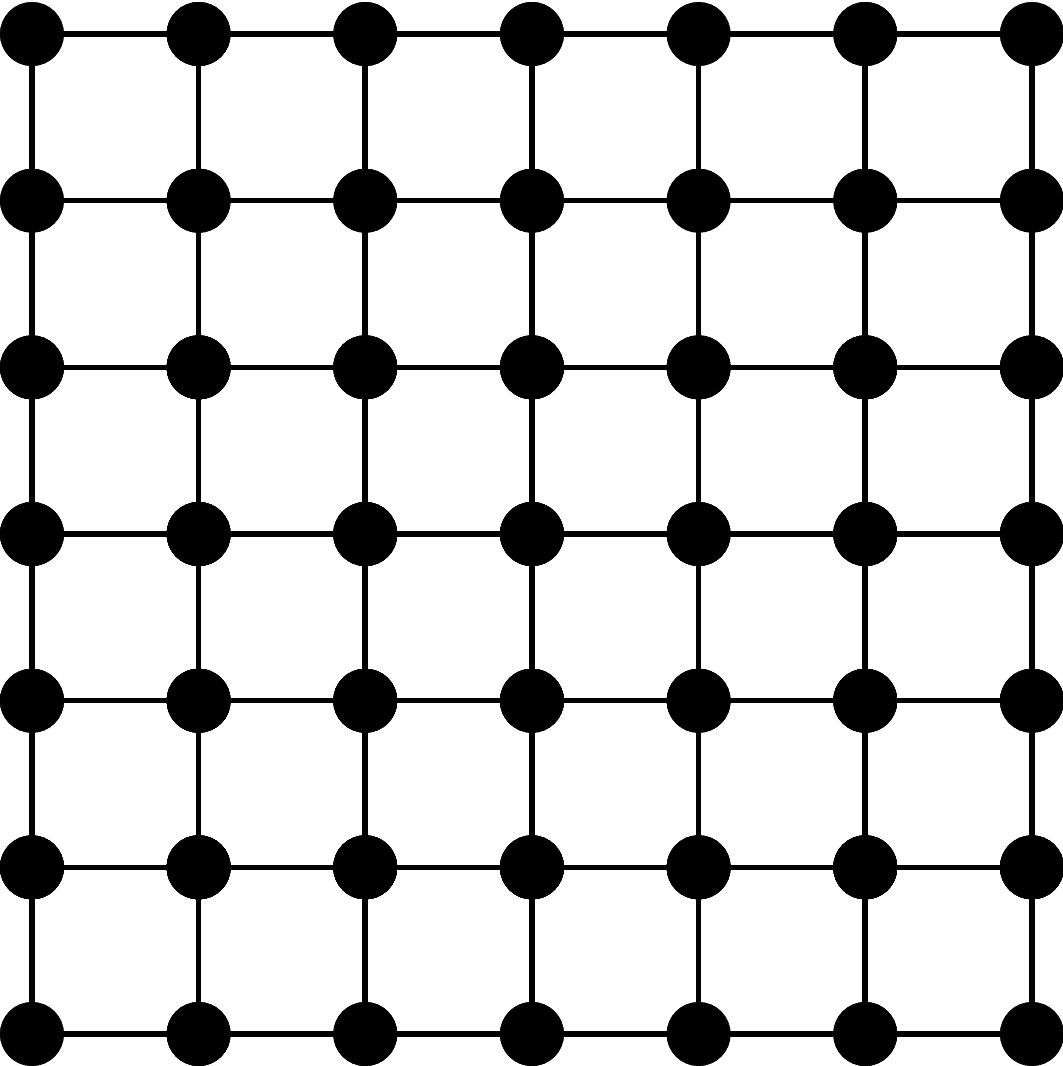}
	\label{fig:7x7grid_input}
	}
	\hfil
\subfigure[Loglikelihood versus number edges]{
	\includegraphics[height=4cm]{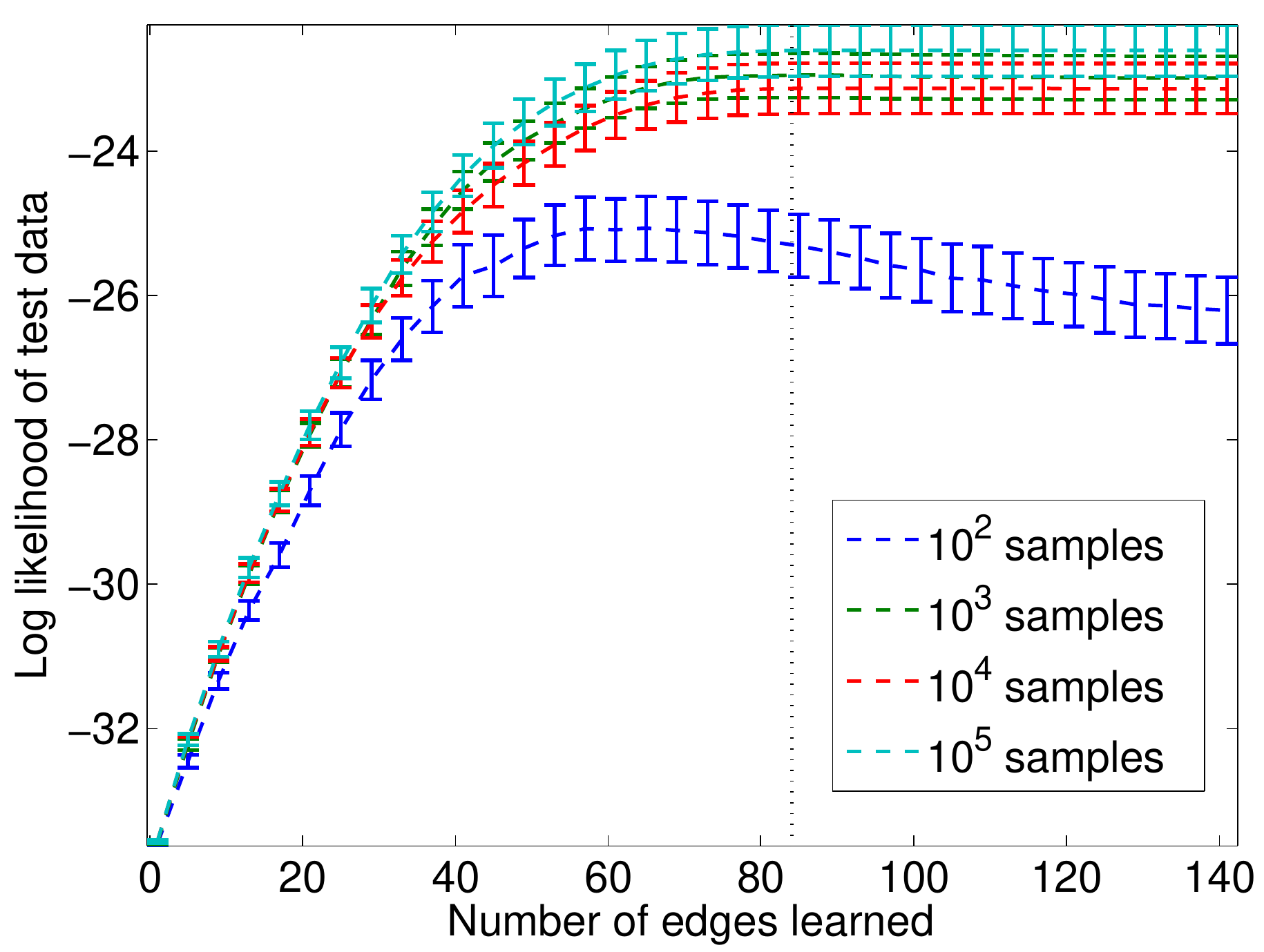}
	\label{fig:7x7grid_loglik_edges}
	}
	\hfil
\subfigure[Comparison of algorithms]{
	\includegraphics[height=4cm]{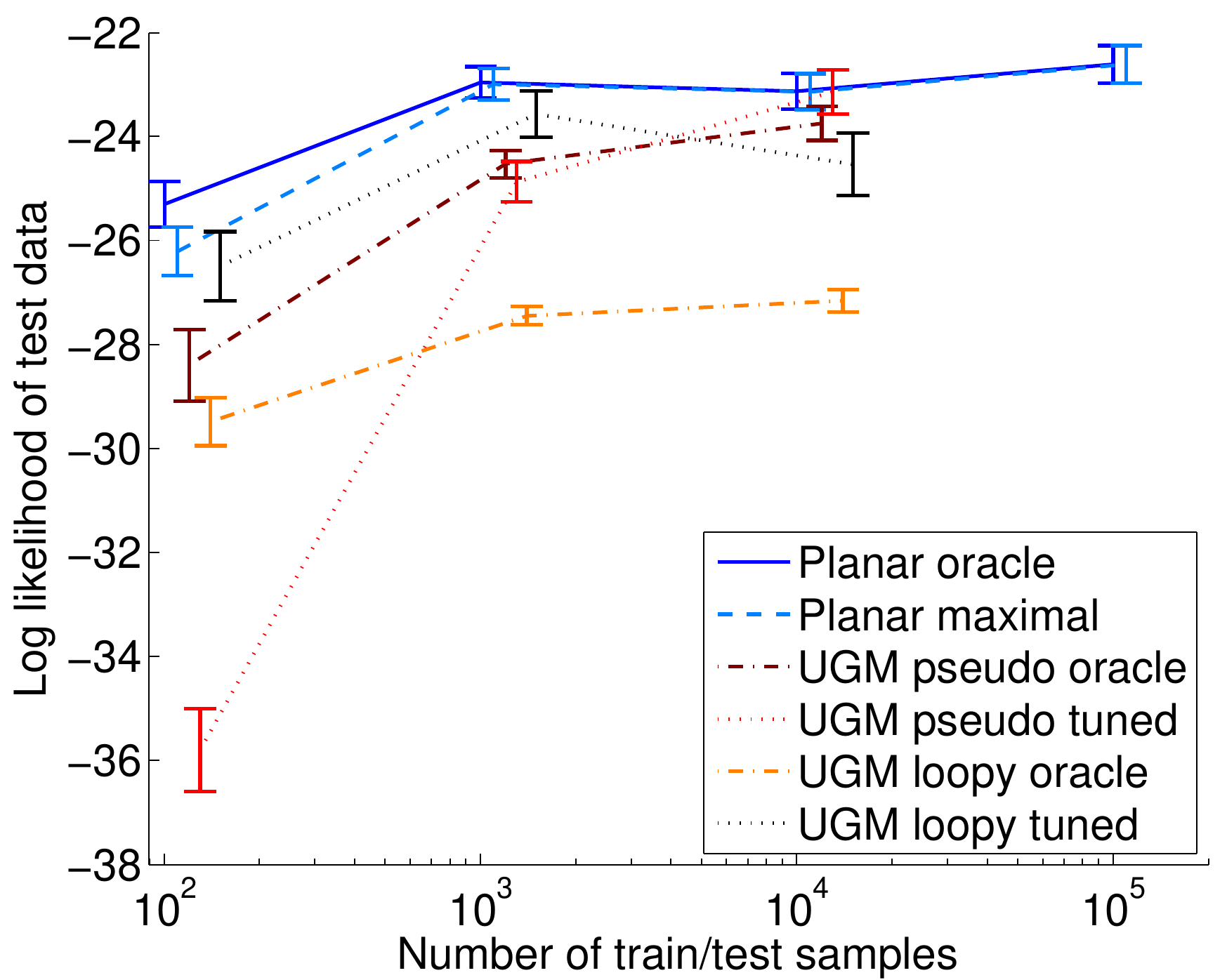}
	\label{fig:7x7grid_loglik}
	}
%
\\
\subfigure[Outer planar]{
	\includegraphics[width=2.2cm]{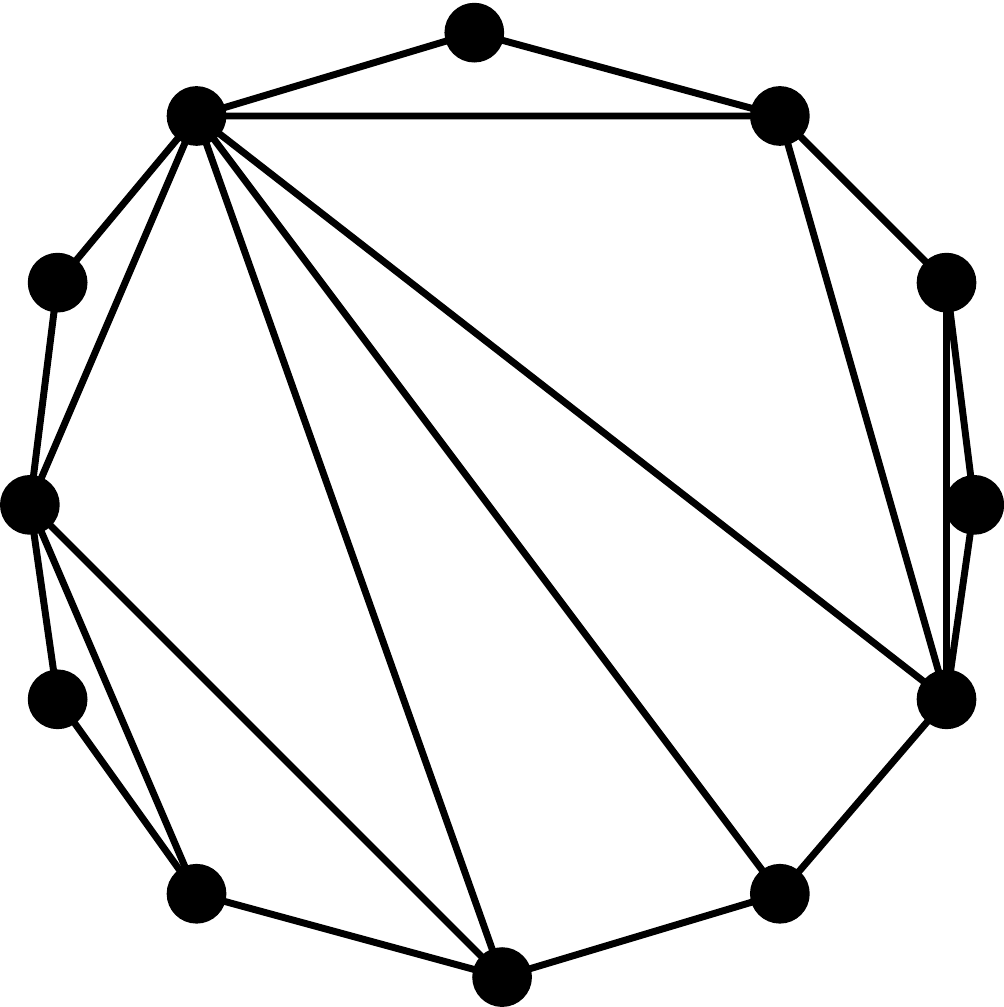}
	\label{fig:outerplanar_input}
	}
	\hfil
\subfigure[Loglikelihood versus number edges]{
	\includegraphics[height=4cm]{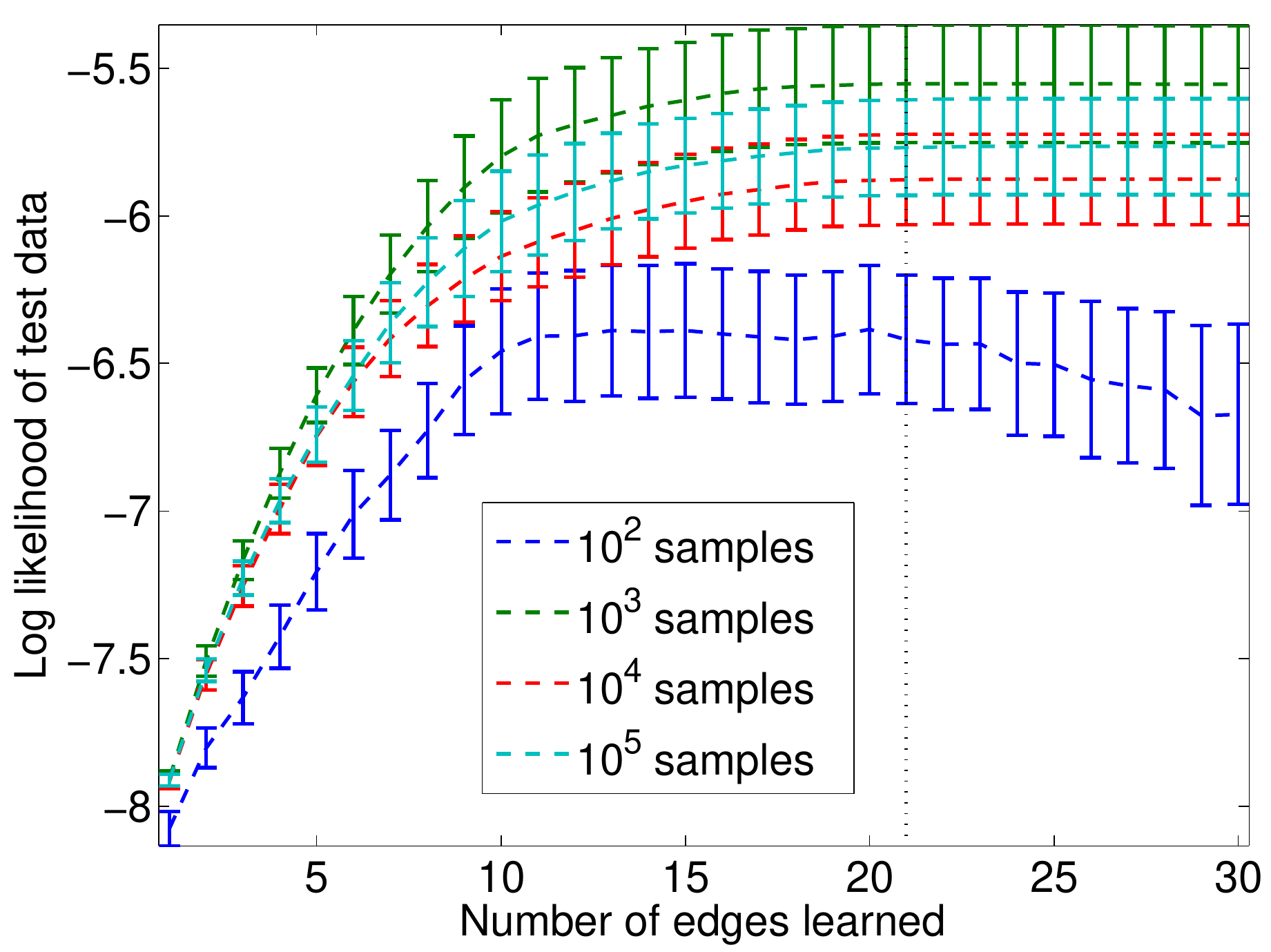}
	\label{fig:outerplanar_loglik_edges}
	}
	\hfil
\subfigure[Comparison of algorithms]{
	\includegraphics[height=4cm]{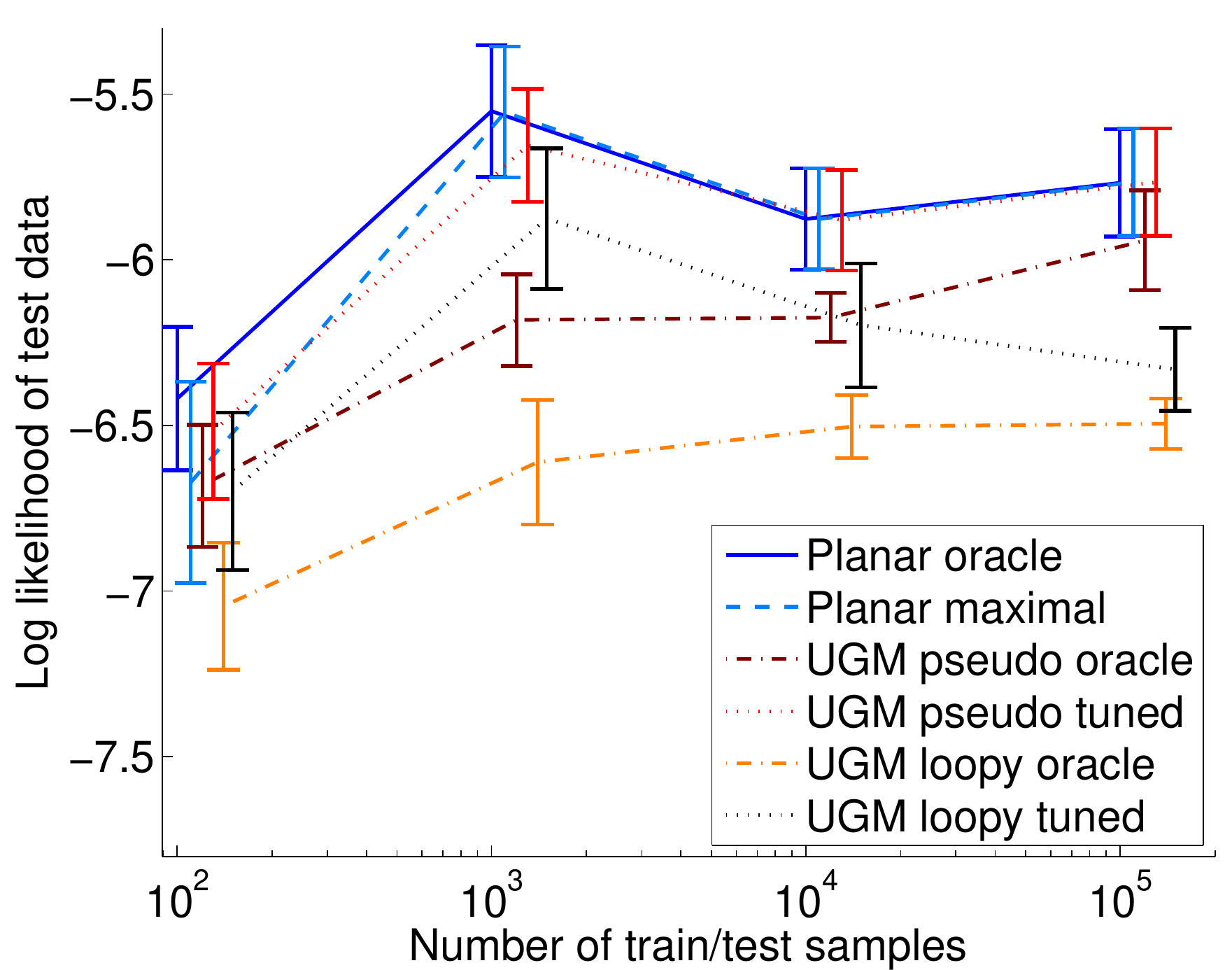}
	\label{fig:outerplanar_loglik_comp}
	}
\vspace{-10pt}
\caption{\small{Results on known models, (\textbf{top} row): $7 \times 7$ grid; and (\textbf{bottom} row): outer planar. \textbf{Left} column (a,d): true graph. \textbf{Middle} column (b,e): likelihood of learned planar graphs as edges are added; and the true number of edges is marked with a vertical dashed line. \textbf{Right} column (c,f): likelihood of test data for various algorithms; x-axis values are perturbed horizontally so that overlapping errorbars are visible.}}
\label{fig:outerplanarResults}
\end{figure*}

We present the results of experiments evaluating our algorithm on known models with
simulated data to evaluate the correctness of the learned models.
We generate two styles of known Ising models: a $7 \times 7$ grid ($n=49$) with zero-field; 
and a $12$-node outer planar model where nodes 
have non-zero mean; shown in Figures~\ref{fig:7x7grid_input} and \ref{fig:outerplanar_input}. The edge parameters are 
chosen uniformly randomly between
$-1$ and $1$ with the condition that the absolute value be greater than a
threshold (chosen to be $0.05$) so as to avoid edges with negligible
interactions. We use Gibbs sampling to obtain samples from this model and
calculate empirical moments from these samples which are then passed as input to
our algorithm. We run 10 trials of randomly 
generated edge parameters and data samples. Though our algorithm can run on graphs with many more nodes, we choose small examples here to
illustrate the result effectively. On the outer planar model, we ensure
that the first moments of all the nodes are satisfied by starting our algorithm with the
auxiliary node connected to all other nodes.

As the planar learning algorithm adds edges to the model, the likelihood of the training data is guaranteed to increase. We assess how adding edges affects the likelihood of out-of-sample test data. Figures~\ref{fig:7x7grid_loglik_edges} and \ref{fig:outerplanar_loglik_edges} demonstrate that likelihood on test sets generally increases as edges are added up to the maximal planar graph. The true number of edges in each synthetic graph is marked with a vertical dotted line. On the smallest datasets ($100$ samples) the out-of-sample performance begins to degrade, a sign of over-fitting the training data; yet the likelihood of the maximal graph is not significantly worse than the best likelihood obtained (with fewer edges).


We also compare against a Markov random field (MRF) learning algorithm for binary data \citep{schmidt2008structure}, as implemented in the undirected graphical model learning Matlab package, UGMLearn\footnote{\url{http://www.cs.ubc.ca/~murphyk/Software/L1CRF}}. UGM is not restricted to learning planar graphs. The objective is optimized via projected gradient descent. We try two versions of the objective function, one using pseudo-likelihood and the other using loopy belief propagation for inference. UGM employs a regularization parameter which we set using two different methods. First, we used the \emph{tuning} method on validation data as detailed in \cite{schmidt2008structure}. That is, we split the data into two parts, train on half the data using 7 different values for the parameter, measure the data likelihood of the other half of the data and vice-versa, then select the parameter value that maximizes the validation data likelihood across both folds. The learned model is trained on the full training data with the tuned regularization parameter value. The second method for setting the regularization parameter we call the \emph{oracle} method, where we select the learned model at the true number of edges, $k$, in our known models. For UGM, we set the regularization parameter via linear search until $k$ edges are learned.

We compare the likelihood of test data from the various learned models in Figures~\ref{fig:7x7grid_loglik} and \ref{fig:outerplanar_loglik_comp}. For comparison, we selected the maximal planar graph that our algorithm learns, \texttt{Planar maximal}; as well as the planar graph learned if the algorithm were stopped when the true number of edges are learned, \texttt{Planar oracle}. We compare against \texttt{UGM pseudo tuned} and \texttt{UGM loopy tuned}, both of which tune the regularization parameter on validation data; but the former uses pseudo-likelihood in learning and the latter uses loopy belief propagation. The tuning method is the most common way of selecting the regularization parameter, but tends to produce relatively dense graphs. For fair comparison, we also show the likelihood of \texttt{UGM pseudo oracle} and \texttt{UGM loopy oracle}; that is, the model with the known true number of edges.

Figures~\ref{fig:7x7grid_loglik} and \ref{fig:outerplanar_loglik_comp} show that our greedy planar
Ising model learning algorithm is at least as accurate and often better 
than the UGM learning algorithms on these
inputs. 
As mentioned earlier, we see that \texttt{Planar maximal} and \texttt{Planar oracle} fit test data nearly equally well.
On the outer planar model, \texttt{UGM pseudo tuned} performs nearly as well as our planar algorithm, yet on the larger grid model it performs quite poorly at the smaller sample sizes. \texttt{UGM loopy tuned} performs more consistently close to our planar algorithm, but it seems that loopy belief propagation performs worse at large sample sizes.

On the largest dataset ($10^5$ samples) of the $7 \times 7$ grid model, UGM was aborted after running for 40 hours without reaching convergence on a single run, and so results are not available.

\section{Application: Modeling Correlations of Senator Voting}

\begin{figure*}[tb]
\vspace{-10pt}
\centering
\subfigure[Learned planar graphical model representing the senator voting pattern]{\includegraphics[width=0.68\textwidth]{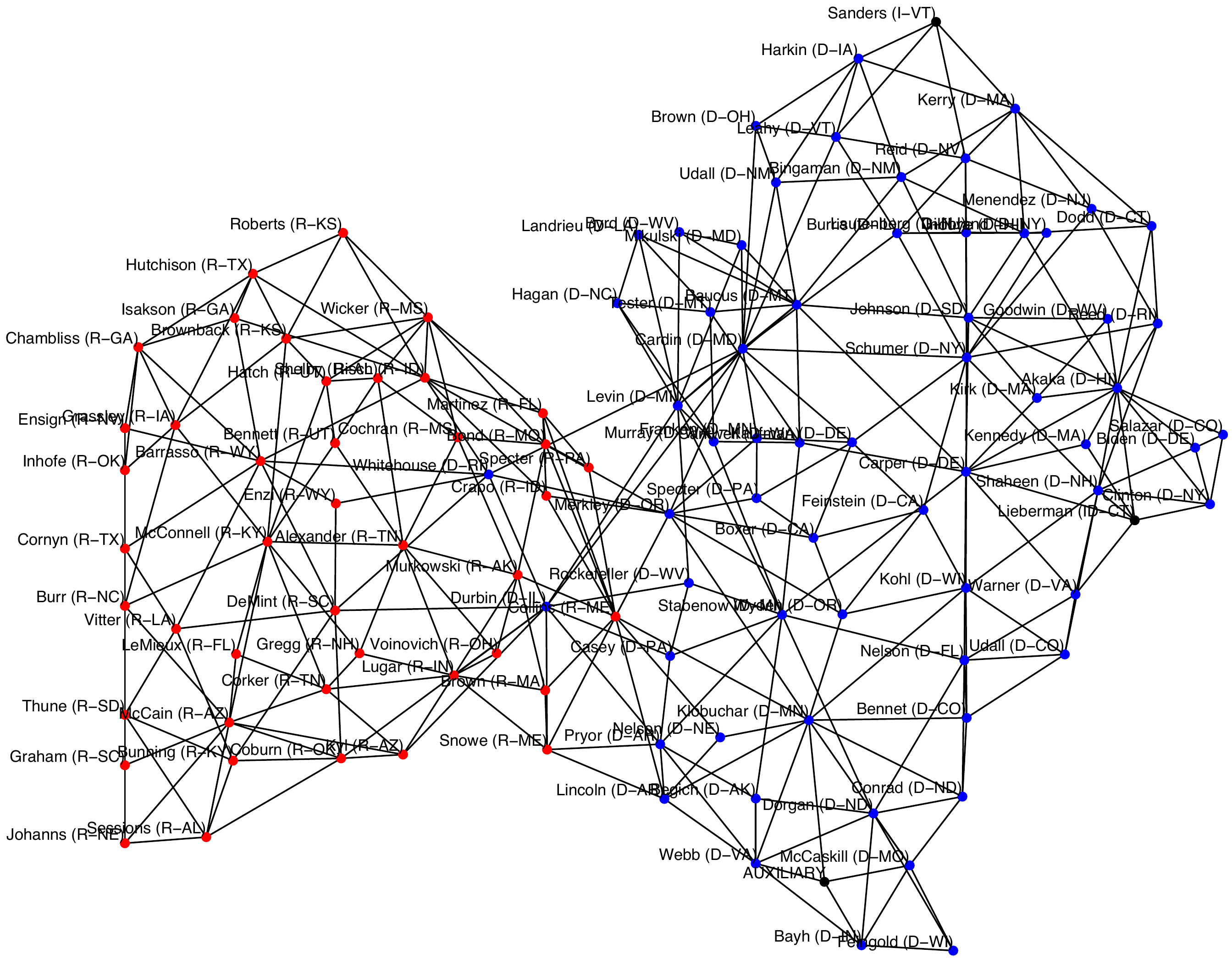}}
\hfil
\subfigure[Comparison of algorithms]{\includegraphics[width=0.3\textwidth]{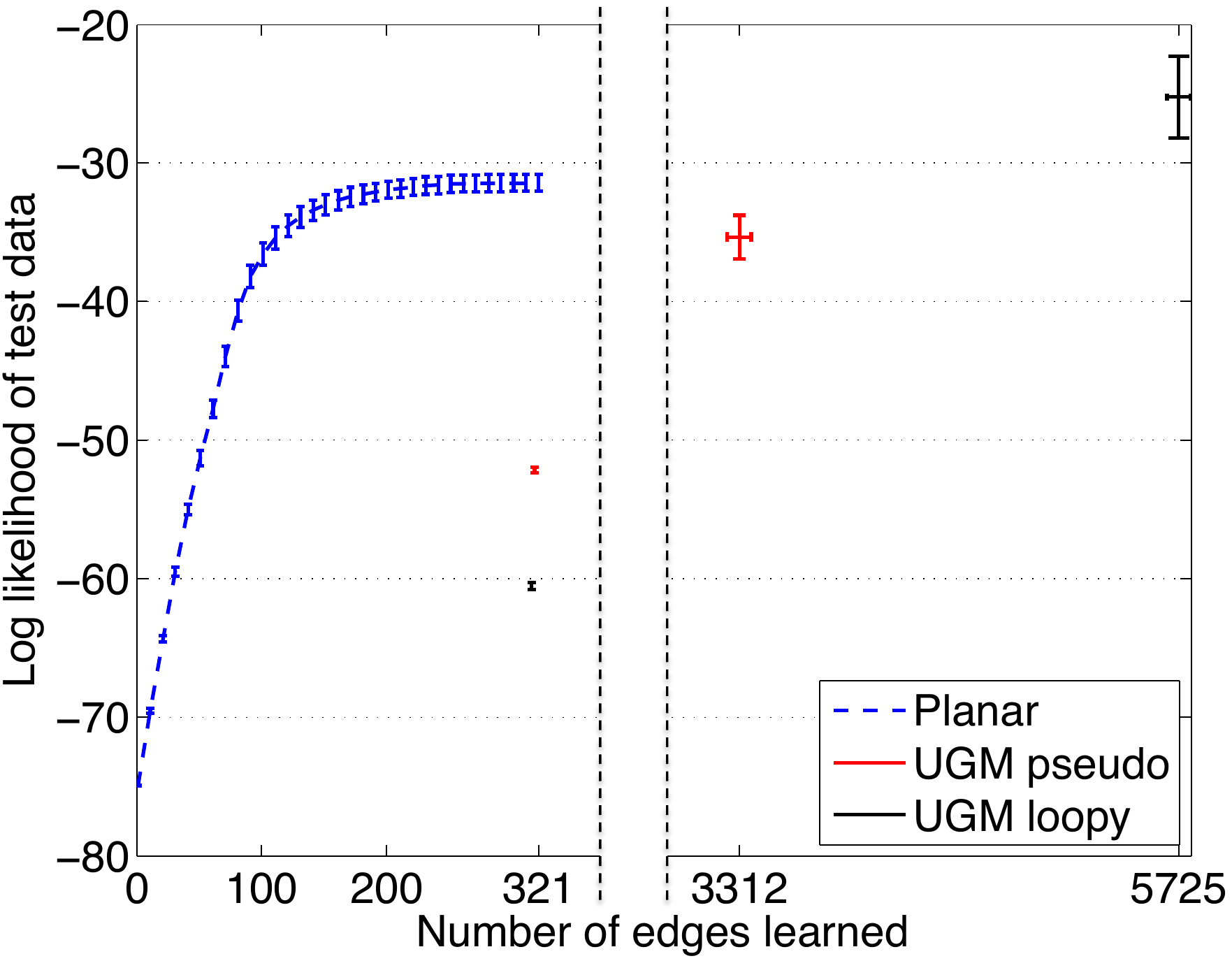}
	\label{fig:senate_loglik}
	}
	\vspace{-10pt}
\caption{\small{Senator voting results. (a) Blue nodes represent Democrats, red nodes
represent Republicans and black nodes represents Independents.
We use a force-directed graph drawing algorithm \citep{fruchterman1991graph}. (b) Likelihood of holdout data versus the number of edges in the learned graph. Note the break in the x-axis, due to tuned UGM learning dense graphs. On the tuned UGM models, we indicate standard error on number of edges learned.}}
\label{fig:senatorgraph}
\end{figure*}


We consider an interesting application of our algorithm to model correlations of senator voting following
\cite{banerjee}.
We use senator voting data from the years 2009 and 2010 to calculate correlations in the voting patterns
among senators. A \textit{Yea} vote is treated as $+1$ and a \textit{Nay} vote is treated as $-1$. We also
treat non-votes as $-1$, but only consider senators who voted in at least $\frac{3}{4}$ of the votes
to limit bias. The data includes $n=108$ variables and 645 samples. To accommodate the non-zero mean data we 
add an auxiliary node and allow the algorithm to select the connections between it and
other nodes. We run a 10-fold cross-validation, training on 90\% of the data and measuring likelihood on the held-out 10\% of data. Figure~\ref{fig:senate_loglik} shows that the likelihood of test data increases as edges are added. We also show the likelihood of cross-validation test data for the \texttt{UGM pseudo} and \texttt{UGM loopy} algorithms for two different methods of choosing the value of the regularization parameter: (1) the value that produces the same number of edges as the maximal planar graph (at 318 edges); and (2) the value selected by tuning with validation data (at a variable number of edges, typically a dense graph). The likelihood of the sparse UGM models are significantly worse than the planar model. Only the \texttt{UGM loopy} algorithm at a very dense (nearly fully connected) graph has better fit to test data.

The maximal planar graph learned from the full dataset, shown in Figure \ref{fig:senatorgraph}, conveys many facts that
are already known to us. For instance, the graph shows Sanders with edges only to Democrats which makes sense because
he caucuses with Democrats. Same is the case with Lieberman. The graph also shows the senate minority leader
McConnell well connected to other Republicans though the same is not true of the senate majority leader Reid. The learned UGM models can be seen in the Supplement, and they show that the non-planar models are qualitatively different, learning one or two densely connected components.

\section{Conclusion and Future Work}

We provide a greedy heuristic to obtain the maximum-likelihood planar
Ising model approximation to a collection of binary random variables with known
pairwise marginals. The algorithm is simple to
implement with the help of known methods for tractable exact inference in planar Ising
models, efficient methods for planarity testing and embedding of planar
graphs. Empirical results of our algorithm on sample data and
on the senate voting record show that it is competitive with arbitrary (non-planar) graph learning.

Many directions for further work are suggested by the
methods and results of this paper. Firstly, we know that the greedy algorithm is
not guaranteed to find the best planar graph. 
In the Supplement, we provide an enlightening counterexample in which the 
combination of the planarity restriction and greedy method prevent the correct model from being learned.
That counterexample suggests strategies one might consider to further refine the estimate.
One strategy would be to allow
the greedy algorithm to prune edges which turn out to be
less important once later edges are added.
It would also be feasible to implement a multi-step greedy look-ahead search technique
for selection of which edge to add (or prune) next.

Another limitation is that our current framework only allows learning planar graphical models
on the set of observed random variables and requires that all
variables are observed in each sample.  One could imagine extensions of our
approach to handle missing samples 
or to try to identify hidden variables that were not seen in the data. This concept
offers another avenue to achieve a better fit to data that is not well-approximated 
by a planar graph among just the set of observed nodes, but might
be well-approximated as the marginal distribution of a planar model with more nodes.


\clearpage
\appendix
\section*{Supplementary Appendix}
\section{Proofs}
\begin{proof}[\textbf{Proposition $1$}]
Let the probability distributions corresponding to $G$ and $\widehat{G}$ be $P$ and $\widehat{P}$ respectively and
the corresponding expectations be $\mathbb{E}$ and $\widehat{\mathbb{E}}$ respectively.
For the partition function, we have that
\begin{equation*}
 \begin{array}{rl}
  \widehat{Z} &= \displaystyle \sum_{x_{\widehat{V}}} \exp\left( \displaystyle \sum_{\{i,j\} \in \widehat{E}} \widehat{\theta}_{ij}x_ix_j
\right) \\
  &= \displaystyle \sum_{x_{\widehat{V}}} \exp\left( x_{n+1}\displaystyle \sum_{i \in V} \theta_{i}x_i+\displaystyle \sum_{\{i,j\} \in E} 
\theta_{ij}x_ix_j\right) \\
  &= \displaystyle \sum_{x_{V}} \exp\left( \displaystyle \sum_{i \in V} \theta_{i}x_i+\displaystyle \sum_{\{i,j\} \in E} \theta_{ij}x_ix_j
\right) \\
  &\;\; + \displaystyle \sum_{x_{V}} \exp\left( -\displaystyle \sum_{i \in V} \theta_{i}x_i+\displaystyle \sum_{\{i,j\} \in E} \theta_{ij}x_ix_j
\right) \\
  &= 2\displaystyle \sum_{x_{V}} \exp\left( \displaystyle \sum_{i \in V} \theta_{i}x_i+\displaystyle \sum_{\{i,j\} \in E} \theta_{ij}x_ix_j
\right) \\
&= 2Z
 \end{array}
\end{equation*}
where the fourth equality follows from the symmetry between $-1$ and $1$ in an Ising model.

For the second part, since $\widehat{P}$ is zero-field, we have that
\begin{equation*}
 \widehat{\mathbb{E}}[x_i] = 0 \; \forall \; i \in \widehat{V}
\end{equation*}
Now consider any $\{i,j\} \in E$. If $x_{n+1}$ is fixed to a value of $1$, 
then the model is the same as original on $V$ and we have
\begin{equation*}
 \widehat{\mathbb{E}}[x_ix_j \mid x_{n+1}=1] =  \mathbb{E}[x_ix_j]\; \forall \; \{i,j\} \in E 
\end{equation*}
By symmetry (between $-1$ and $1$) in the model, the same is true for 
$x_{n+1}=-1$ and so we have
\begin{equation*}
  \begin{array}{rl}
    &\widehat{\mathbb{E}}[x_ix_j]\\
    &=\widehat{\mathbb{E}}[x_ix_j \mid x_{n+1}=1]\widehat{P}(x_{n+1}=1)\\
    &\;+ \widehat{\mathbb{E}}[x_ix_j \mid x_{n+1}=-1]\widehat{P}(x_{n+1}=-1)\\
    &= \mathbb{E}[x_ix_j]
  \end{array}
\end{equation*}
Fixing $x_{n+1}$ to a value of $1$, we have
\begin{equation*}
 \widehat{\mathbb{E}}[x_i \mid x_{n+1}=1] =  \mathbb{E}[x_i]\; \forall \; i \in V
\end{equation*}
and by symmetry
\begin{equation*}
 \widehat{\mathbb{E}}[x_i \mid x_{n+1}=-1] =  -\mathbb{E}[x_i]\; \forall \; i \in V
\end{equation*}
Combining the two equations above, we have
\begin{equation*}
  \begin{array}{rl}
    &\widehat{\mathbb{E}}[x_ix_{n+1}]\\
    &=\widehat{\mathbb{E}}[x_i \mid x_{n+1}=1]\widehat{P}(x_{n+1}=1)\\
    &\;+ \widehat{\mathbb{E}}[-x_i \mid x_{n+1}=-1]\widehat{P}(x_{n+1}=-1)\\
    &= \mathbb{E}[x_i]
  \end{array}
\end{equation*}
\end{proof}

\begin{proof}[\textbf{Proposition $2$}]
 From Theorem $1$, we see that the log partition function can be written as
\begin{equation*}
 \Phi(\theta) = n \log 2 + \displaystyle \sum_{\{i,j\}\in E} \log \cosh \theta_{ij} + \frac{1}{2} \log \det(I-AD)
\end{equation*}
where $A$ and $D$ are as given in Theorem $1$. 
For the derivatives, we have
\begin{equation*}
\begin{array}{rl}
  \frac{\partial\Phi(\theta)}{\partial \theta_{ij}} &= \tanh \theta_{ij} + \frac{1}{2} \mbox{Tr}\left((I-AD)^{-1} \frac{\partial(I-AD)}{\partial 
\theta_{ij}}\right)\\
	&= \tanh \theta_{ij} - \frac{1}{2} \mbox{Tr}\left((I-AD)^{-1} A D'_{ij}\right)\\
	&= w_{ij} - \frac{1}{2} (1-w_{ij})^2 \left( S_{ij,ij}+S_{ji,ji}\right)
\end{array}
\end{equation*}
where $D'_{ij}$ is the derivative of the matrix $D$ with respect to $\theta_{ij}$.
The first equality follows from chain rule and the fact that $\nabla K = K^{-1}$ for any matrix $K$. Please refer to
\cite{boyd} for details.

For the Hessian, we have
\begin{equation*}
 \begin{array}{rl}
  \frac{\partial^2\Phi(\theta)}{\partial \theta_{ij}^2} &= \frac{1}{Z(\theta)}\frac{\partial^2 Z(\theta)}{\partial \theta_{ij}^2} - \frac{1}
{Z(\theta)^2}\left( \frac{\partial Z(\theta)}{\partial \theta_{ij}} \right)^2 \\
	  &= 1 -\mu_{ij}^2
 \end{array}
\end{equation*}
For $\{i,j\}\neq \{k,l\}$, following \cite{boyd}, we have
\begin{equation*}
\begin{array}{rl}
 \frac{\partial^2\Phi(\theta)}{\partial \theta_{ij}\partial \theta_{kl}}
   &= -\frac{1}{2}\mbox{Tr}\left( SD'_{ij}SD'_{kl}\right) \\
   &= -\frac{1}{2} (1-w_{ij}^2) \left( S_{ij,kl}S_{kl,ij}+S_{ji,kl}S_{kl,ji} \right. \\
   &\;\;\;\;\left. +S_{ij,lk}S_{lk,ij}+S_{ji,lk}S_{lk,ji}\right) (1-w_{kl}^2)
\end{array}
\end{equation*}
On the other hand, we also have
\begin{equation*}
 \begin{array}{rl}
  T_{ij,kl} &= e_{ij}^T (I+P) (S\circ S^T) (I+P) e_{kl} \\
	   &= (e_{ij} + e_{ji})^T (S\circ S^T) (e_{kl} + e_{lk}) \\
	   &= (S\circ S^T)_{ij,kl} + (S\circ S^T)_{ij,lk} \\
	   &\;\;\;+ (S\circ S^T)_{ji,kl} + (S\circ S^T)_{ji,lk} \\
	   &= S_{ij,kl} S_{kl,ij}+S_{ji,kl}S_{kl,ji} \\
	   &\;\;\;+S_{ij,lk}S_{lk,ij}+S_{ji,lk}S_{lk,ji}
 \end{array}
\end{equation*}
where $e_{ij}$ is the unit vector with $1$ in the $ij^{\mbox{th}}$ position and $0$ everywhere else.
Using the above two equations, we obtain
\begin{equation*}
 H_{ij,kl} = -\frac{1}{2} (1-w_{ij}^2) T_{ij,kl} (1-w_{kl}^2)
\end{equation*}
\end{proof}

\begin{proof}[\textbf{Proposition $4$}]
The proof follows from the following steps of inequalities.
\begin{equation*}
 \begin{array}{rl}
  D(P,P_G) &= D(P,P_{G+ij}) + D(P_{G+ij},P_G) \\
	   &= D(P,P_{G+ij}) + \\
	   &\;\;\;D(P_{G+ij}(x_i,x_j),P_G(x_i,x_j)) + \\
	   &\;\;\;D(P_{G+ij}(x_{V-ij}),P_G(x_{V-ij})) \\
	   &\geq D(P,P_{G+ij}) + \\
	   &\;\;\;D(P_{G+ij}(x_i,x_j),P_G(x_i,x_j)) + \\
	   &\geq D(P,P_{G+ij}) + \\
	   &\;\;\;D(P(x_i,x_j),P_G(x_i,x_j))
 \end{array}
\end{equation*}
where the first step follows from the Pythagorean law of information projection \citep{amari},
the second step follows from the conditional rule of relative entropy \citep{cover},
the third step follows from the information inequality \citep{cover} and finally
the fourth step follows from the property of information projection to $G+ij$ \citep{wainwright}.
\end{proof}

\section{Experiments: Counter Example}

\begin{figure}[tb]
\vspace{-10pt}
\centering
\subfigure[Counter example original]
{
\includegraphics[width=0.25\textwidth]{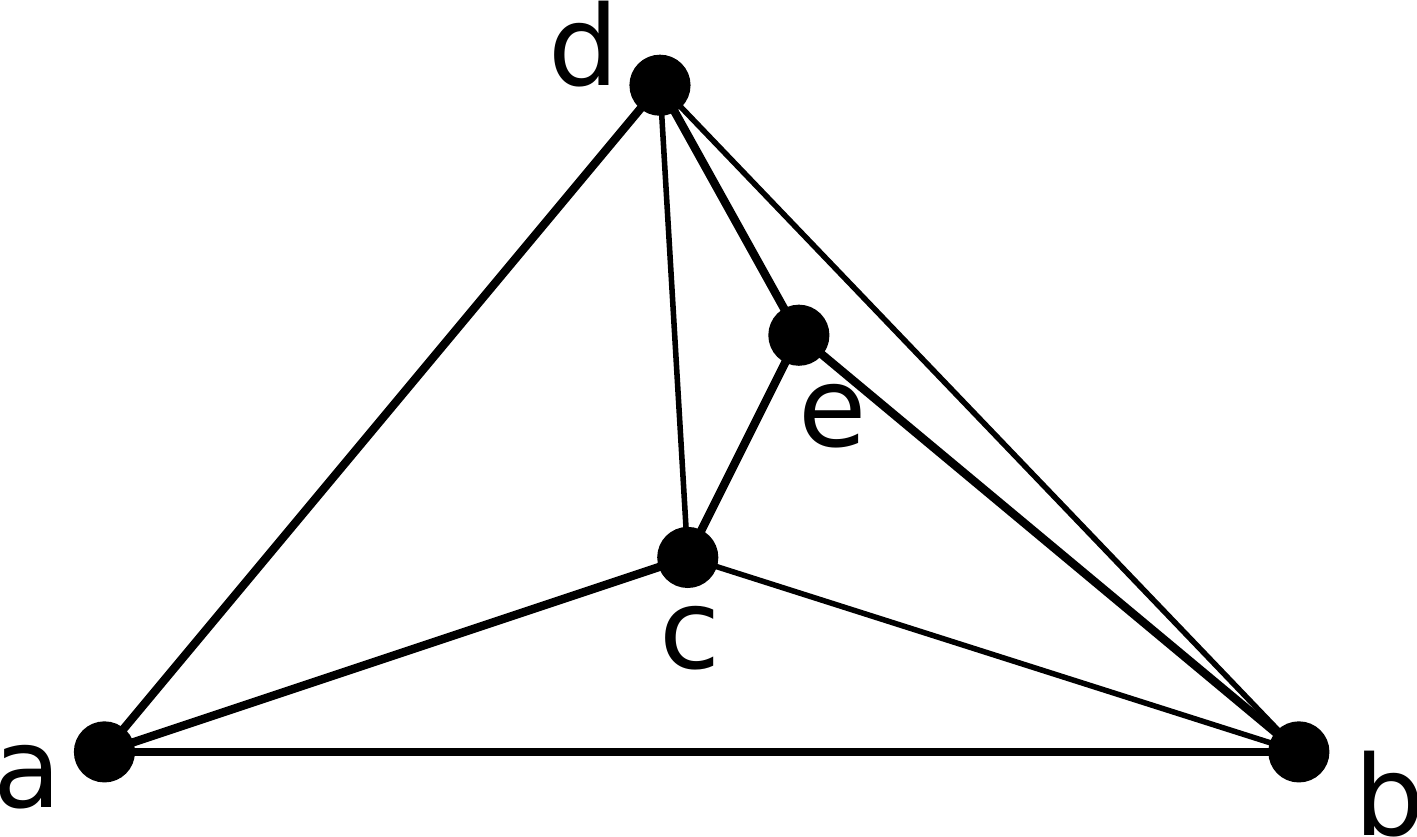}
\label{fig:counterexample_input}
}
\hfil
\subfigure[Recovered model]
{
\includegraphics[width=0.25\textwidth]{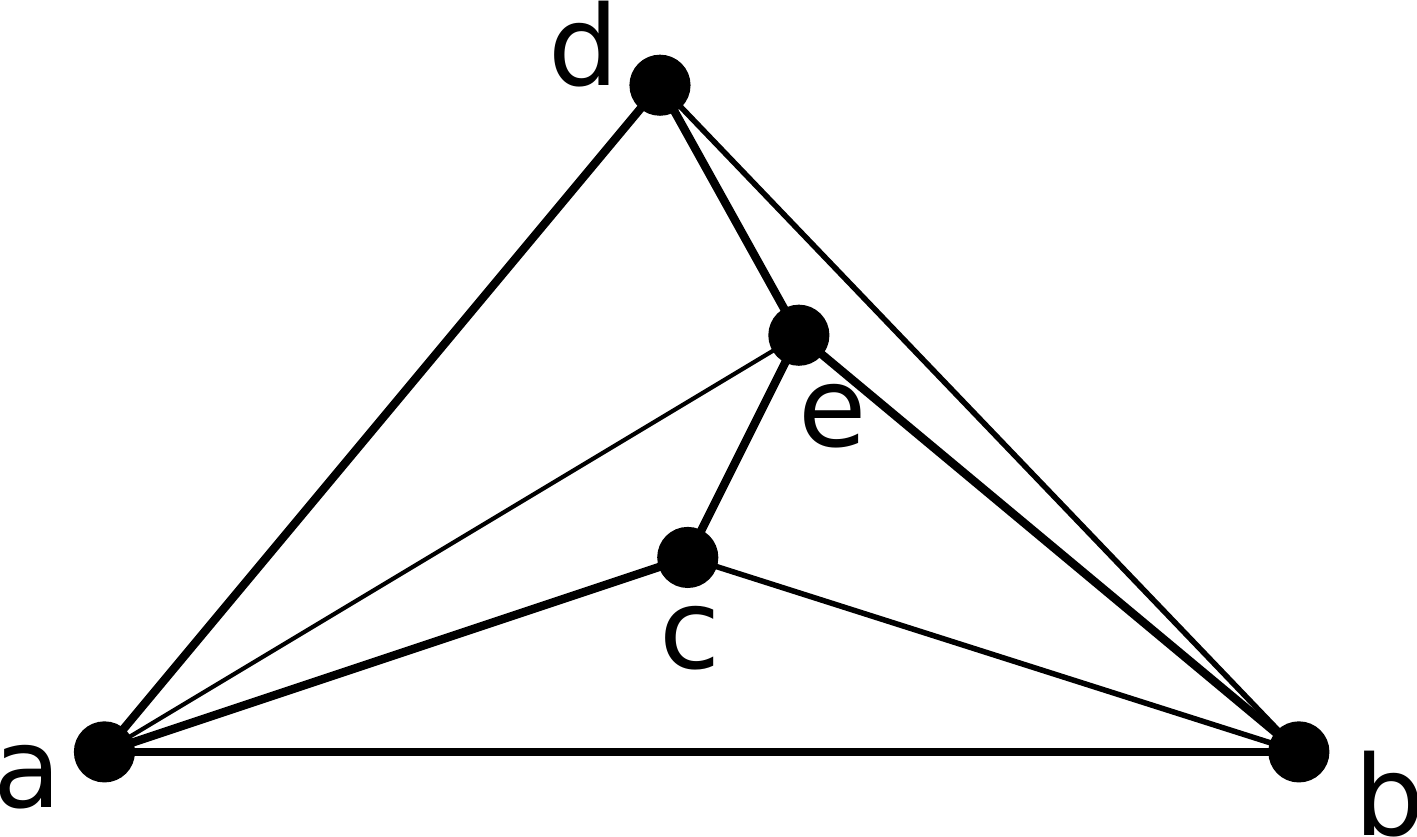}
\label{fig:counterexample_output}
}
\caption{Example graphical models. (a) Counter example. (b) The recovered graphical model
has one spurious edge $\{a,e\}$ and one missing edge $\{c,d\}$. }
\label{fig:counterexample}
\end{figure}

The result presented in Figure~\ref{fig:counterexample} illustrates the fact that our 
algorithm does not always recover the exact
structure even when the underlying graph is planar and the algorithm is given exact moments as
inputs. This counterexample gives insight into how the greedy algorithm works.
The basic idea is that graphical models can have nodes which are not neighbors but are more correlated
than some other nodes which are neighbors. If the spurious edges corresponding to these highly correlated nodes
are added early on in the algorithm, then the actual edges may have to be left out because of the planarity
restriction.

We define a zero-field Ising model on the graph in Figure
\ref{fig:counterexample_input} with the edge parameters as follows:
$\theta_{bc}=\theta_{cd}=\theta_{bd}=0.1$ and $\theta_{ij}=1$ for all the other
edges. Figure \ref{fig:counterexample_input} shows the edge parameters in the
graph pictorially using the intensity of the edges - higher the intensity of an
edge, higher the corresponding edge parameter. With these edge parameters,
 the correlation between nodes $a$ and $e$ is greater than the
correlation between any other pair of nodes. This leads to the edge between $a$
and $e$ to be the first edge added in the algorithm. However, since $K5$ (the
complete graph on $5$ nodes) is not planar, one of the actual edges
is missed in the output graph as shown in Figure \ref{fig:counterexample_output}.

\section{Example Application: UGM Learned Models}

\begin{figure*}[tb]
\centering
\includegraphics[width=\textwidth]{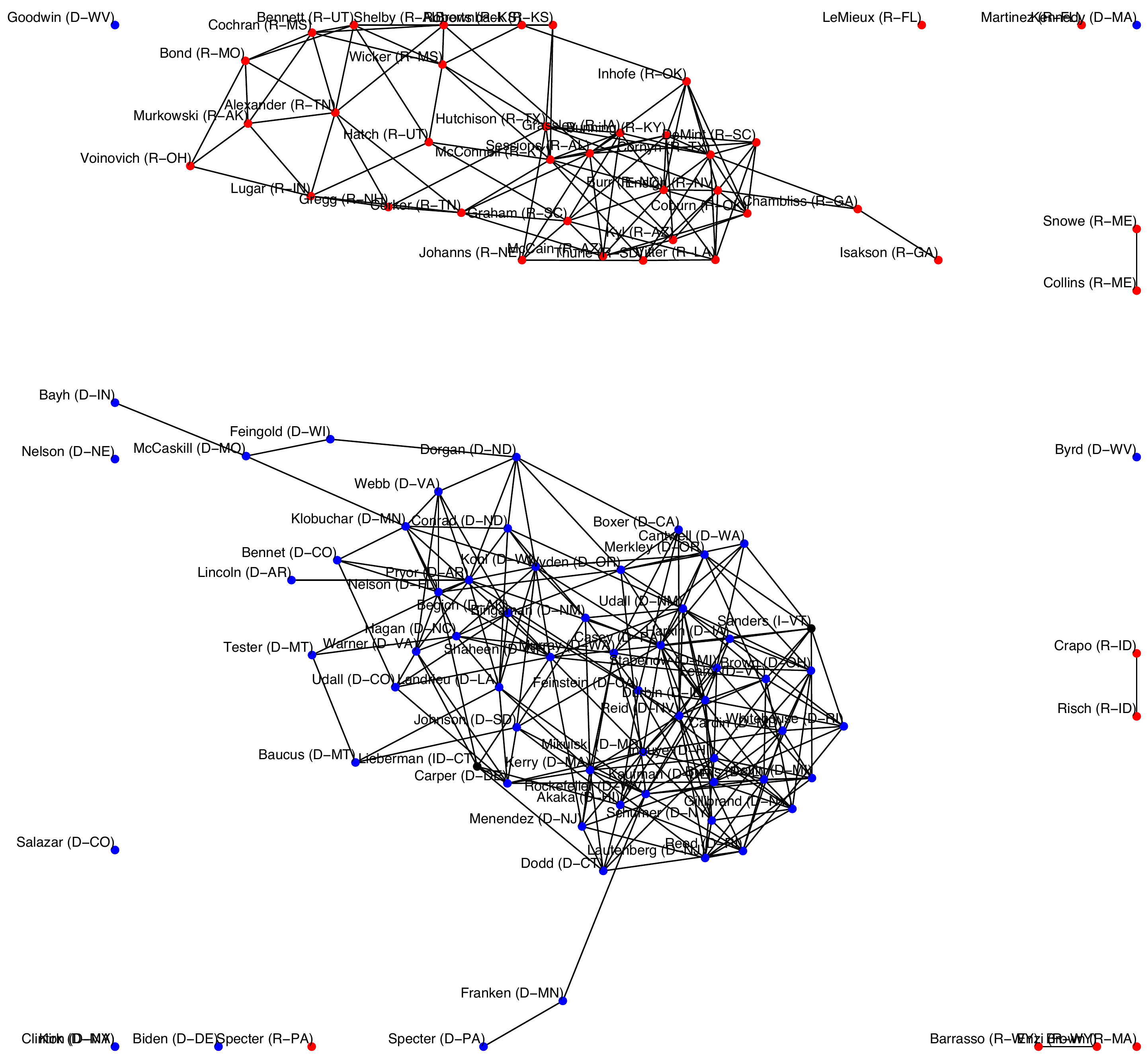}
\caption{Senate voting graph learned by \texttt{UGM pseudo} with 318 edges. Blue nodes represent Democrats, red nodes represent Republicans and black nodes represent Independents.}
\label{fig:senate_ugm}
\end{figure*}

\begin{figure*}[tb]
\centering
\includegraphics[width=0.6\textwidth]{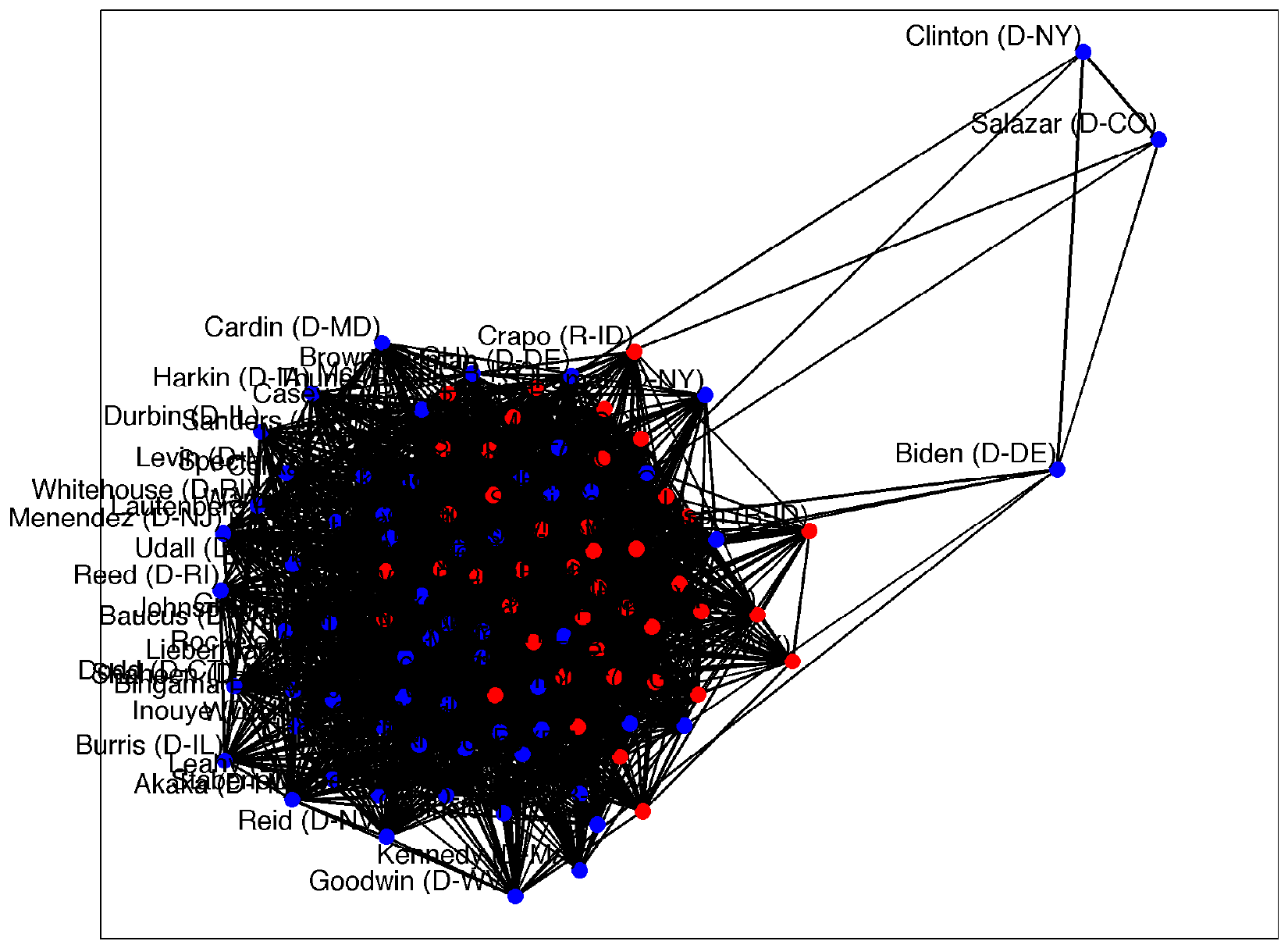}
\caption{Senate voting graph learned by \texttt{UGM pseudo tuned}. Blue nodes represent Democrats, red nodes represent Republicans and black nodes represent Independents.}
\label{fig:senate_ugm_tune}
\end{figure*}

\begin{figure*}[tb]
\centering
\includegraphics[width=\textwidth]{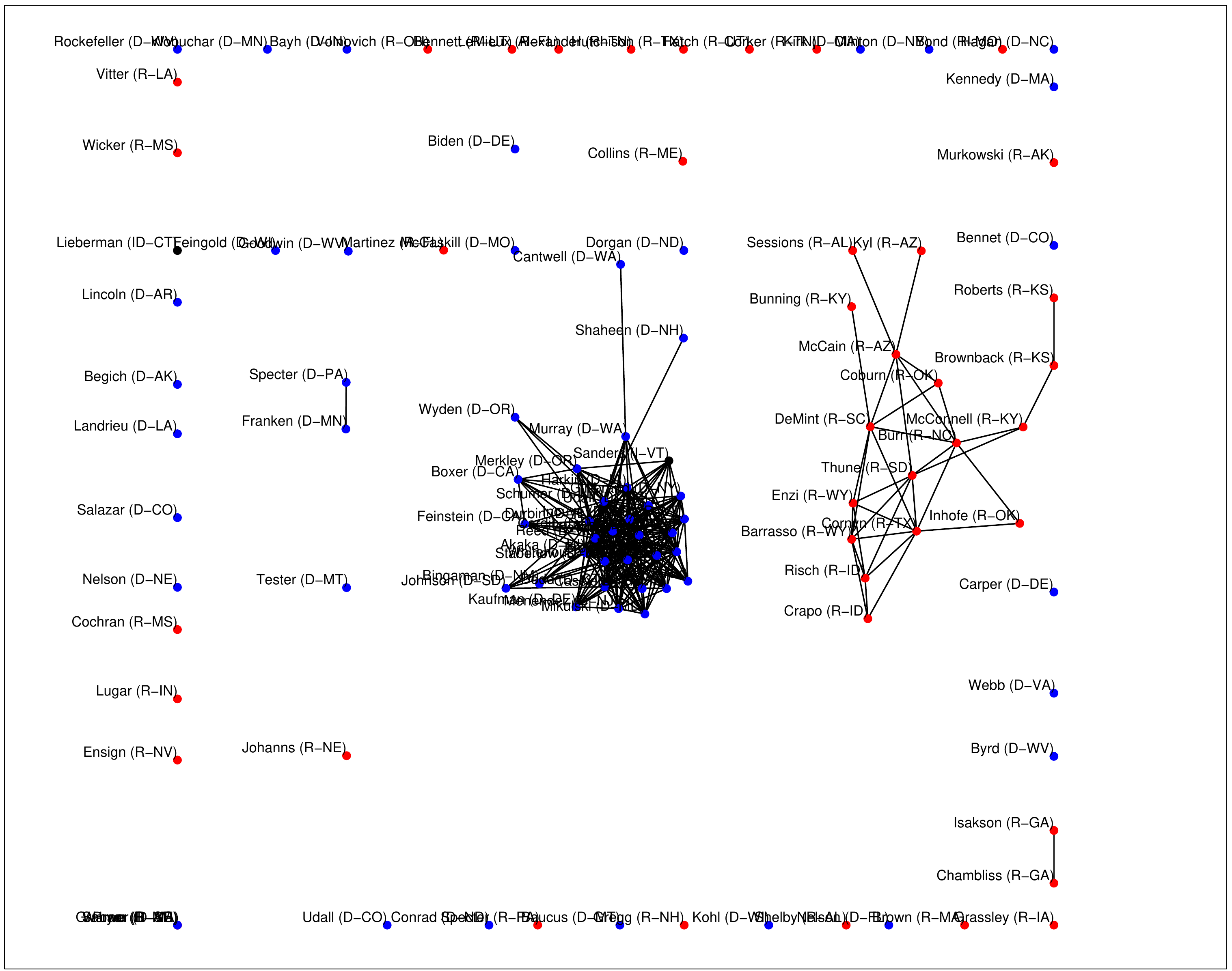}
\caption{Senate voting graph learned by \texttt{UGM loopy} with 318 edges. Blue nodes represent Democrats, red nodes represent Republicans and black nodes represent Independents.}
\label{fig:senate_ugmloop}
\end{figure*}

For comparison to our planar learning algorithm, we provide the results of using the UGM MRF learning algorithm on the senate voting data. For all figures, we use a force-directed graph drawing algorithm \citep{fruchterman1991graph}. Figure~\ref{fig:senate_ugm} presents the graph learned using pseudolikelihood, \texttt{UGM pseudo}, from the full dataset with the regularization parameter set to obtain the same number of edges as learned in the planar case (318 edges). Figure~\ref{fig:senate_ugm_tune} presents the graph learned using pseudolikelihood, \texttt{UGM pseudo tuned}, from the full dataset after selecting the regularization parameter from cross-validation tuning. Figure~\ref{fig:senate_ugmloop} presents the graph learned using loopy belief propagation, \texttt{UGM loopy}, from the full dataset with the regularization parameter set to obtain 318 edges. The graph learned using \texttt{UGM loopy tuned} is not displayed because it is a nearly fully-connected graph providing no visual information.

\clearpage
\bibliographystyle{abbrvnat}
\begin{small}
\bibliography{planar}
\end{small}

\end{document}